%% file: LegoWalker.tex
\documentclass[10pt,a4paper]{article}
\pdfoutput=1
\usepackage[T1]{fontenc}
\usepackage[margin=1in,includehead,headheight=11pt,headsep=11pt,columnsep=2pc]{geometry}
\usepackage{mathptmx}
\usepackage{amsmath,amssymb}
\usepackage{microtype}
\usepackage{graphicx}
\usepackage{float}
\usepackage{caption}
\usepackage{subcaption}
\usepackage{booktabs}
\usepackage[ruled,vlined,linesnumbered]{algorithm2e}
\usepackage{placeins}
\usepackage{fancyhdr}
\usepackage{enumitem}
\usepackage{cite}
\usepackage{xcolor}
\usepackage{url}
\usepackage{listings}
\usepackage[most]{tcolorbox}
\usepackage[colorlinks=true,linkcolor=blue,citecolor=blue,urlcolor=blue]{hyperref}
\hypersetup{
 pdftitle={Make it Walk: Computational Design of LEGO Walkers},
 pdfauthor={Siqi Li, Fancheng Gao, Chi-Wing Fu, Peng Song},
 pdfsubject={Computational design of LEGO walkers},
 pdfkeywords={LEGO walker, walking automaton, computational design, mechanism design}
}
\setlist{leftmargin=1.5em,labelsep=0.5em,itemsep=0pt plus 0.1pt,
 parsep=0pt,topsep=0pt plus 1pt minus 0.5pt,partopsep=0pt}
\makeatletter
\renewcommand{\section}{\@startsection{section}{1}{\z@}
 {-11pt plus -5.5pt minus -1pt}{0.001pt}
 {\raggedright\normalfont\normalsize\bfseries}}
\renewcommand{\subsection}{\@startsection{subsection}{2}{\z@}
 {-11pt plus -5.5pt minus -1pt}{0.001pt}
 {\raggedright\normalfont\normalsize\bfseries}}
\renewcommand{\subsubsection}{\@startsection{subsubsection}{3}{\z@}
 {-5.5pt plus -5.5pt minus -1pt}{0.001pt}
 {\raggedright\normalfont\normalsize\bfseries}}
\renewcommand{\paragraph}{\@startsection{paragraph}{4}{\z@}
 {5.5pt plus 2.25pt minus 1pt}{-0.5em}{\normalfont\normalsize\bfseries}}
\renewcommand{\@seccntformat}[1]{\csname the#1\endcsname.\hspace{0.5em}}
\renewcommand{\@maketitle}{%
 \newpage\null\vskip 20pt
 \begingroup
 \centering\setlength{\parskip}{0pt}
 {\fontsize{17.28}{20}\selectfont\bfseries\@title\par}
 \vskip 25pt
 {\normalsize\begin{tabular}[t]{c}\@author\end{tabular}\par}
 \vskip 18pt
 \endgroup
}
\makeatother

\makeatletter

\renewcommand{\@biblabel}[1]{[#1]\hspace{0.5em}}
\makeatother

\renewenvironment{abstract}{%
 \par\vspace{2pt}\hrule height 0.25pt\vspace{3pt}
 \begin{list}{}{\setlength{\leftmargin}{2pc}\setlength{\rightmargin}{2pc}
  \setlength{\topsep}{0pt}\setlength{\partopsep}{0pt}\setlength{\parsep}{0pt}}
 \item[]\normalsize\textbf{Abstract}\par\itshape\noindent
}{\end{list}\vspace{3pt}\hrule height 0.25pt\par}

\title{Make it Walk: Computational Design of LEGO\textsuperscript{\textnormal{\textregistered}} Walkers}

\author{
  Siqi Li\textsuperscript{1}\quad
  Fancheng Gao\textsuperscript{1}\quad
  Chi-Wing Fu\textsuperscript{2}\quad
  Peng Song\textsuperscript{1}\\[4pt]
  {\normalfont\small\textsuperscript{1}Singapore University of Technology and Design}\\
  {\normalfont\small\textsuperscript{2}The Chinese University of Hong Kong}\\[4pt]
}
\date{}

\begin{document}
\raggedbottom
\twocolumn[{
\begin{minipage}{\textwidth}
\maketitle

\begingroup
\setlength{\intextsep}{0pt}
  \captionsetup{hypcap=false}
  \centering
  \includegraphics[width=0.96\linewidth]{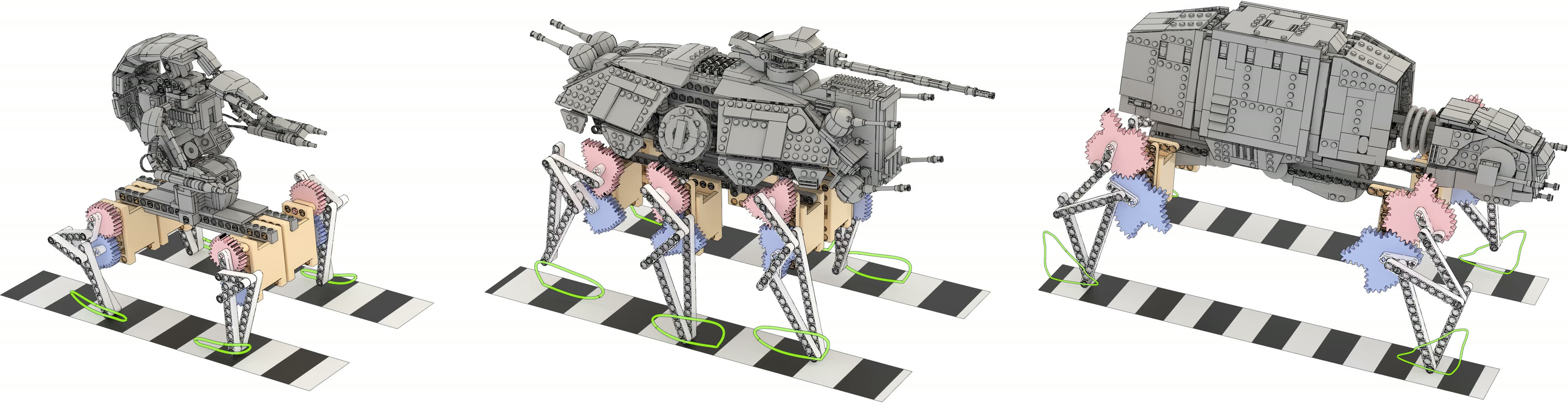}
  \vspace*{-2mm}
  \captionof{figure}{
    LEGO walkers designed by our method.
    Note their different body frames and walking gaits (in green).
    Our method computes and optimizes the leg mechanisms, constructing the
    LEGO Technic assemblies (in white) and the non-circular gears
    (in pink and blue), which together deliver various walking gaits while enabling stable walking.
  }
  \label{fig:teaser}
\par
\endgroup

\input{sec0-abstract}
\vspace{12pt}
\end{minipage}
}]
\thispagestyle{empty}

\input{sec1-introduction}

\input{sec2-relatedwork}

\input{sec3-problem}
\input{sec4-modeling}

\input{sec5-design}

\input{sec6-result}

\input{sec7-conclusion}

\bibliographystyle{alpha-doi}
\bibliography{Ref_LegoWalker}

\end{document}

%% file: sec0-abstract.tex
\begin{abstract}
	LEGO walkers are walking automata constructed from modular LEGO Technic bricks. They exhibit intriguing locomotion and are also highly reconfigurable through the modular 
    Technic bricks, making them widely applicable in education, recreation, and robotics research. However, existing LEGO walkers are designed mostly by reusing existing single-actuator walking mechanisms.
	Although the locomotion is stable, their fixed walking gaits limit their adaptability and versatility. This paper presents the first computational solution capable of designing LEGO walkers that not only walk stably, but also support a variety of walking gaits.
    Our idea is twofold. First, we control the foot motion trajectory of each leg mechanically via a specifically-designed gear-linkage mechanism, in which we incorporate a pair of 3D-printed non-circular gears to work with a LEGO Technic mechanism. Second, we drive each leg by an independent motor and synchronize the timings of multiple leg motions via a simple yet adaptable control algorithm, enabling stable and versatile locomotion. Importantly, we take this modeling as a foundation to co-design the morphology and control of the LEGO walkers to meet the target locomotion and appearance. Leveraging our approach, we created a variety of LEGO walkers, demonstrating various walking gaits and appearances, and fabricated some of them to show their locomotion capability in the real world.

\end{abstract}

%% file: sec1-introduction.tex
\section{Introduction}
\label{sec:intro}
Walking automata are mechanical systems designed to produce legged locomotion. 
The LEGO Technic system provides an accessible and modular platform for rapid prototyping walking automata, namely {\em LEGO walkers\/}. 
These walkers deliver intriguing 
locomotion behaviors and have been widely used in 
recreational robots~\cite{Dittert-2024-BricksRL} and robotics research. 
In addition, they serve as effective STEM education tools, nurturing hands-on exploration of the gait principles and stable creeping locomotion.

\begin{figure}[t]
	\centering
	\includegraphics[width=0.99\columnwidth]{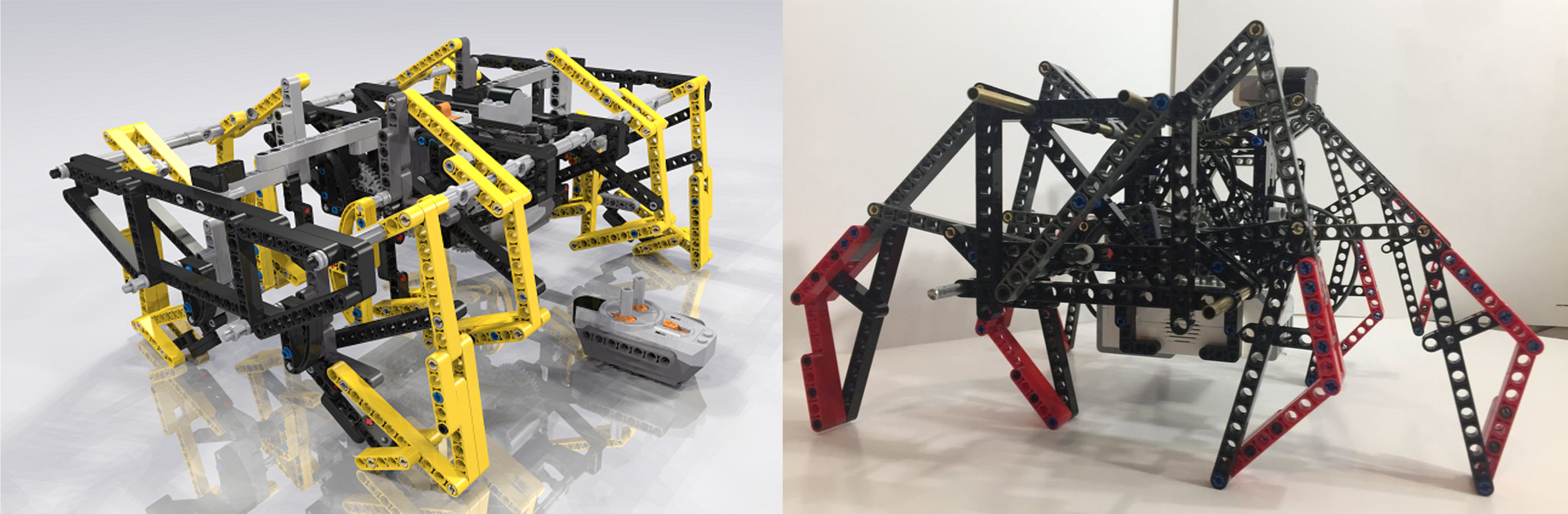}
	\vspace{-2mm}
	\caption{Existing LEGO walkers~\cite{Walker-Youtube} often rely on bulky and complex mechanisms, yet delivering only a fixed walking gait.}
	\label{fig:walker_from_internet}
	\vspace{-2mm}
\end{figure}

Existing works create LEGO walkers largely by reusing existing linkage-based walking mechanisms such as Klann linkage~\cite{Klann-2001-Walking} and Jansen's linkage~\cite{Jansen-2008-Strandbeests}; see Figure~\ref{fig:walker_from_internet} for examples.
Once assembled, such designs are able to walk stably, driven by a single actuator. 
Yet, they are specifically constrained to deliver few fixed walking gaits pre-defined in existing walking mechanisms, limiting their versatility and expressiveness.

In this work, we present a new computational solution to create LEGO walkers that can deliver a variety of walking gaits while walking stably.
This problem is non-trivial for two reasons.
First, we cannot rely on existing walking mechanisms, since they generally produce a fixed walking gait.
Hence, the first challenge asks for modeling the morphology of the LEGO walker, including which mechanical parts to use, how to connect them, and also how to actuate them.
The second challenge lies in the vast design space of our problem.
To achieve stable walking with varying gaits, we need to determine not only the geometry and layout of the mechanical parts but also the control and actuation of these parts.

To address these challenges, our key idea is to decouple the {\em trajectory} and {\em timing} control of each walking leg, simplifying the walker's morphology and control.
At the leg level, we propose to control the foot motion trajectory of each leg {\em mechanically} via a specifically-designed gear-linkage mechanism and incorporate a pair of 3D-printed non-circular gears to work with the LEGO Technic mechanism.
In this way, we can enlarge the design space of the walking gait.
At the walker level, we propose to drive each leg by an independent motor and synchronize the timings of multiple leg motions via {\em electronically} controlling the actuation of the driving motors.
By this means, each leg is modeled with its own gear-linkage mechanism, so our LEGO walker can feature a simple and lightweight morphology for delivering stable walking with varying gaits; see Figures~\ref{fig:teaser} and~\ref{fig:RobotComponents} for examples.

To design LEGO walkers, we 
only need to specify the desired walking gait (including the foot trajectory and footfall pattern) and provide a body frame of the walker. 
Then, our computational approach co-designs the morphology and control of the LEGO walker to meet the target walking gait and walker appearance.
In particular, our approach first constructs and optimizes the leg mechanism to approximate the target foot trajectory, then generates the leg layout to match the given body frame, and optimizes the control on the leg actuations for stable walking guided by the gait theory.
At the end, our approach generates all the information necessary for fabricating the LEGO walker, including 
the LEGO Technic brick assemblies, 
the shape of the non-circular gears for 3D printing, 
the assembly instructions, and 
the control for the servo motors.
We validate our approach through the design and physical realization of several LEGO-based quadruped and hexapod walkers, demonstrating the practicality and versatility of our approach. 

In this work, we make the following contributions:
\begin{itemize}[leftmargin=*]
    \item
	We exploit a new class of LEGO walkers that integrates mechanical control (for generating leg trajectories) and electrical control (for synchronizing leg motions) for realizing a variety of walking gaits, %
    where the core is a modular and lightweight leg mechanism modeled as a five-bar linkage with two non-circular gears.
	\vspace*{1mm}
	\item
	We propose an optimization-based approach, capable of efficiently co-designing the morphology and control of LEGO walkers, satisfying the target walking gait and walker appearance, while ensuring stable walking and fabricability.
\end{itemize}

%% file: sec2-relatedwork.tex
\begin{figure}[!t]
	\vspace{-5pt}
	\centering
	\includegraphics[width=0.99\columnwidth]{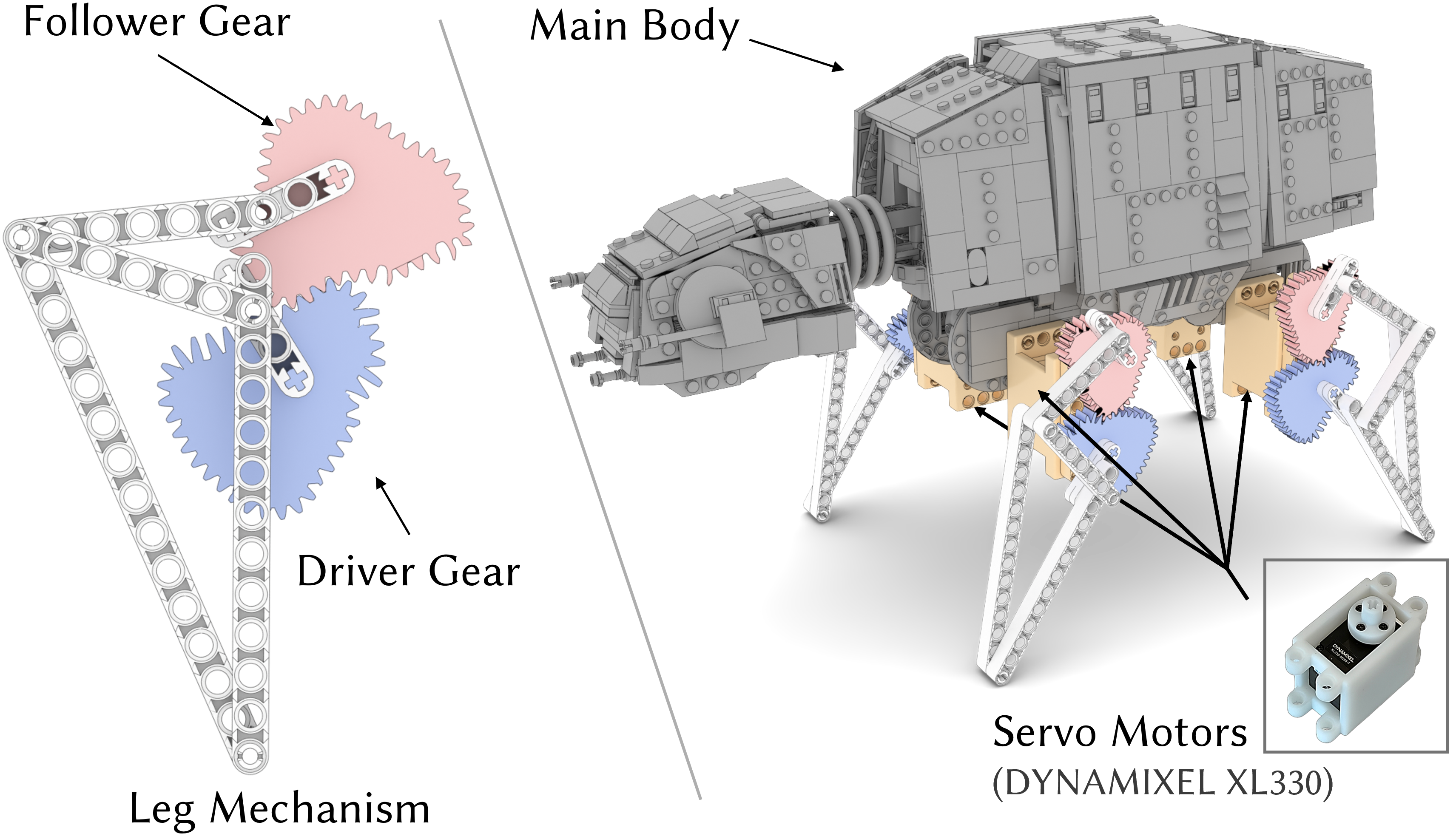}
	\vspace{-1mm}
	\caption{Illustration of our LEGO walker.
		Left: our leg mechanism comprises a five-bar linkage of LEGO Technic bricks together with a pair of non-circular gears.
		Right: each leg is attached to the main body and actuated by an independent motor.}
	\vspace{-2mm}
	\label{fig:RobotComponents}
\end{figure}

\section{Related Work}
\label{sec:related}

\paragraph*{Walking automaton design.}
With the advance and growing popularity of digital fabrication, the computational design and fabrication of personalized walking automata has attracted immense interest from the graphics community.
A number of computational methods have been developed to design automata that perform intriguing walking motion \emph{in mid air}, 
including 
mechanical characters~\cite{Coros-2013-MechanicalCharacter},
mechanical figures~\cite{Ceylan-2013-MechanicalAutomata},
linkage-based characters~\cite{Thomaszewski-2014-Linkage},
kinetic wire characters~\cite{Xu-2018-Bend-It}, and
compliant mechanical systems~\cite{Tang-2020-CompliantPeriodicMotion}.

Beyond displaying walking motion, other research has explored the design of automata capable of walking stably on the ground.
Bharaj et al.~\cite{Bharaj-2015-WalkingAutomata} developed a computational tool for designing personalized walking automata, where the automata are first interactively created by users from a library of predefined template linkage mechanisms and then optimized to achieve stable walking on the ground. 
Cheng et al.~\cite{Cheng-2021-3DCamDesign} designed a six-leg automaton with a fixed walking gait based on a composite 3D cam-follower mechanism.

\begin{figure*}[!t]
	\center
	\includegraphics[width=0.99\textwidth]{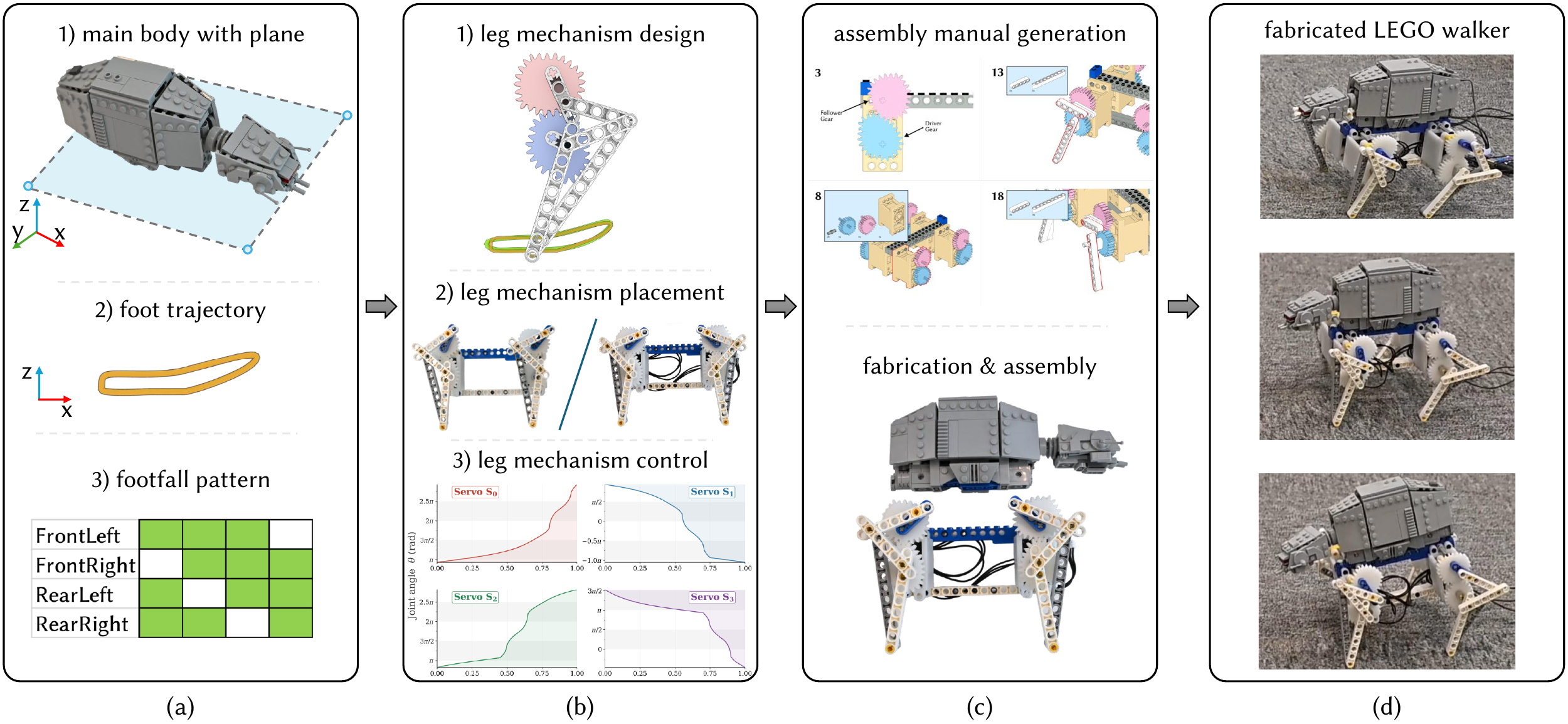}
	\vspace{-2mm}
	\caption{
		Overview of our approach. %
		(a) %
        The input %
        includes a main body with a plane for leg placement (in blue), a target foot trajectory, and a footfall pattern. (b) %
        Our approach %
        automatically designs a compatible leg mechanism and optimizes the placement of the legs and their actuation control. 
        (c) Our approach also generates
        a LEGO-style assembly manual, providing %
        instructions %
        to aid the assembly of the walker.
        (d) Our fabricated LEGO walker can walk stably, %
        following the target gait.
	}
	\label{fig:PipelineOverview}
	\vspace{-2mm}
\end{figure*}

\paragraph*{Walking robot design.}
A walking robot is a legged robotic system that moves across the ground, where each leg is typically built of an open kinematic chain with multiple actuated joints.
Several computational approaches have been developed for designing different kinds of walking robots, including
robotic creatures~\cite{Megaro-2015-RoboticCreature, Geilinger-2018-Skaterbots} and 
skinned quad-robots~\cite{Feng-2021-SkinnedQuad-Robots}.
In particular, Zhao et al.~\cite{Zhao-2020-RoboGrammar} presented a computational approach called RoboGrammar for generating optimized structures of walking robots to traverse given terrains. 

Although walking robots provide versatile locomotion capabilities by supporting a wide variety of walking gaits, they need a large number of actuators to drive every robot joint. Hence, they are often expensive to fabricate and complex to control. %
Due to low structural stiffness and relatively large joint clearance, LEGO Technic systems are rarely used for building walking robots. In this work, we study the computational design and fabrication of LEGO walkers, which are inexpensive to fabricate and easy to control, while supporting stable walking with varying gaits.

\paragraph*{LEGO Technic design.}
LEGO Technic is an advanced building system for creating complex and functional models built of LEGO Technic bricks such as
beams, axles, pins, and gears.
Existing research works~\cite{Luo-2015-Legolization, Zhou-2019-LegoArchitecture,
Pun-2025-Generating, Ge-2025-Lego, Xu-2025-LegoACE}
on creating LEGO models mainly focus on designing static structures with non-Technic bricks. %
Regarding LEGO Technic, Xu et al.~\cite{Xu-2019-LEGOTechnic} introduced a method to compute and construct LEGO Technic models that follow user-specified sketches. 
To the best of our knowledge, we are not aware of any research work on constructing LEGO Technic walkers.

%% file: sec3-problem.tex
\section{Problem Statement and Overview}
\label{sec:problem}
We address the problem of computationally designing multi-legged LEGO walkers. 
Our approach takes the following %
inputs, as illustrated in Figure~\ref{fig:PipelineOverview}(a):
\begin{itemize}[leftmargin=*]
	\vspace{-1mm}
    \item 
    {\em Body frame} \
    includes a desired LEGO structure that serves as the main body of the walker, and 
    a plane specified on the body for attaching the leg mechanisms to be designed.
	\vspace{1mm}
	\item 
	{\em Walking gait} \
    includes
    a target foot locus and a footfall pattern. 
	The foot locus describes the desired trajectory of the foot for a leg to follow, %
    whereas the footfall pattern specifies the timing and sequence of the leg-ground contact of each leg. 
\end{itemize}

Given these inputs, we aim to create a fabrication-ready LEGO walker that satisfies the requirements for the walking gait and walker appearance.
The output of our approach includes: 
\begin{itemize}[leftmargin=*]
    \vspace{-1mm}
    \item 
    {\em Geometry and fabrication package of the leg mechanisms,} \
    including 
    an assembly of LEGO Technic bricks, profiles of non-circular gears, and step-by-step assembly instructions;
    \vspace{1mm}
	\item 
    {\em Layout of the leg mechanisms,} \
	defined by the pose of each leg mechanism relative to the main body; and
    \vspace{1mm}
    \item 
    {\em Control of the leg mechanisms,} \
    defined by the coordinated, time parameterized position trajectories for the servo motors required to achieve the target gait while maintaining balance.
\end{itemize}

\paragraph*{Overview of our approach.}
Figure~\ref{fig:PipelineOverview} gives an overview of our approach.
Section~\ref{sec:modeling} models the morphology and actuation of LEGO walkers, where each leg is modeled as a gear-linkage mechanism driven by an independent actuator.
Section~\ref{sec:design} formulates an optimization to co-design the morphology and control of LEGO walkers and presents a method to effectively solve the optimization to find a desired LEGO walker design.

%% file: sec4-modeling.tex
\section{Modeling LEGO Walkers} 
\label{sec:modeling}

In this section, we first introduce how we model the leg mechanism as a geared five-bar linkage mechanism, and then introduce the actuation of each leg mechanism and the overall walker model. Finally, we introduce stable walking gaits that guide the locomotion of LEGO walkers.

\subsection{Modeling Leg Mechanisms}

\begin{figure}[t]
    \includegraphics[width=0.99\columnwidth]{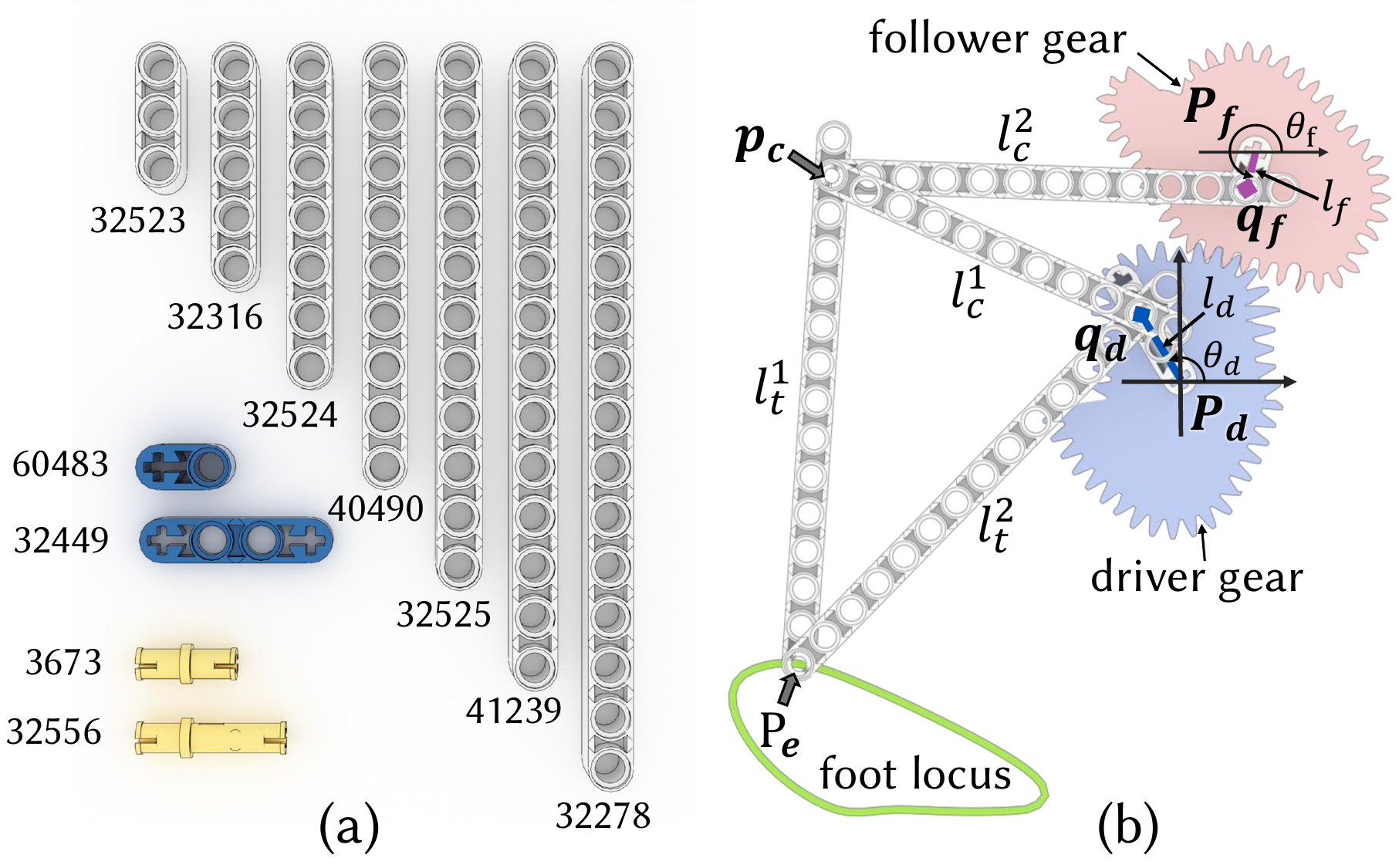}
    \vspace{-2mm}
    \caption{
    Our leg mechanism.
    (a) Our employed LEGO Technic bricks, each with a unique ID.
    The gray beams are the main links in our leg mechanism; the blue beams with axle holes are for connecting with gears; and the yellow pins are revolute joints for connecting the links while allowing relative motion between them. (b) The schematic of our leg mechanism, which is a five-bar linkage %
    with a pair of non-circular gears. 
    The rotation centers of the two gears are fixed relative to the body frame.
    }
    \label{fig:leg_mechanism}
    \vspace{-2mm}
\end{figure}

Among various leg mechanism designs, we chose the five-bar linkage as the basic structure for our leg design for three reasons. First, it has two degrees of freedom (DOFs), sufficient for generating diverse foot trajectories. Second, its topology is relatively simple, making the mechanism compact and efficient. Third, linkage mechanisms are well suited for LEGO implementation, as they can be readily built using standard LEGO beams and pins. 
To simplify the control, we further reduce the mechanism to a single degree of freedom, coupling two driving joints of the five-bar linkage with a pair of non-circular gears. 
This allows the leg to be driven by a single actuator while still producing nontrivial foot trajectories.
Figure~\ref{fig:leg_mechanism} illustrates our leg mechanism and its components. 

\paragraph*{2-DOF five-bar linkage.}
The five-bar linkage contains two fixed base joints $\mathbf{p}_d, \mathbf{p}_f$ (where subscript d means driver and f means follower), two input links of lengths $l_d$ and $l_f$, and two coupler links of lengths $l_c^1$ and $l_c^2$ that connect to a shared coupler point $\mathbf{p}_c$. In addition, two passive links of lengths $l_t^1$ and $l_t^2$ form a rigid triangular foot structure attached to the coupler. The geometry of the linkage mechanism is parameterized by the link lengths $\mathbf{L} \in \mathbb{Z}^6$ and the base offset $\mathbf{p}_f \in \mathbb{Z}^2$, with $\mathbf{p}_d$ fixed as the origin of the leg's local frame. All parameters are defined over integer coordinates to respect the discrete geometric constraints of LEGO Technic bricks and their connections. %

\noindent
{\em Forward Kinematics.} \
The 2-DOF five-bar linkage is driven by two input links with angles $(\theta_d, \theta_f)$.
Given input angles $(\theta_d, \theta_f)$, the positions of the intermediate joints are
\begin{equation}
    \begin{aligned}
    \mathbf{q}_d &= \mathbf{p}_d + l_d
    \begin{bmatrix}\cos \theta_d \ \sin \theta_d\end{bmatrix}, \\
    \mathbf{q}_f &= \mathbf{p}_f + l_f
    \begin{bmatrix}\cos \theta_f \ \sin \theta_f\end{bmatrix}.
    \end{aligned}
\end{equation}
The coupler point $\mathbf{p}_c$ is obtained by solving 
\begin{equation}
    |\mathbf{p}_c - \mathbf{q}_d| = l_c^1
    \ \mathrm{and} \
    |\mathbf{p}_c - \mathbf{q}_f| = l_c^2.
\end{equation}
The passive links together with $\mathbf{p}_c$ form a rigid triangle. 
Thus, the end-effector (foot) position $\mathbf{P}_e$ can be expressed as a fixed offset from $\mathbf{p}_c$ under the triangle's configuration. The forward kinematics of the five-bar linkage can be expressed as a function of the input angles and leg geometry:
\begin{equation}
    \mathbf{P}_{e} = \mathbf{f}(\theta_d, \theta_f; \mathbf{L}, \mathbf{p}_f).
    \label{eq:forward_kinematics}
\end{equation}

Singularities occur when the two circles $(\mathbf{q}_d, l_c^1)$ and $(\mathbf{q}_f, l_c^2)$ become tangent, i.e.,
\begin{equation}
|\mathbf{q}_d - \mathbf{q}_f| = |l_c^1 \pm l_c^2|,
\end{equation}
at which point the mechanism loses mobility and the forward kinematics become ill-conditioned. These configurations lead to unstable motion and should be avoided in the design process.

\paragraph*{Non-circular gear actuation.}
To realize leg motion using only a single actuator for the five-bar linkage mechanism, we couple the two input links with a pair of non-circular gears. The driver gear is actuated by a servo motor and the follower gear is actuated by the driver gear. The two input links are connected to the two gears, respectively. For simplicity, we first rotate the frame of the two gears to align with the center line between $\mathbf{p}_d$ and $\mathbf{p}_f$, and we denote the driver and follower angles in this frame as $\phi_d$ and $\phi_f$, respectively. 

The links are rigidly connected to the gears, so we can express the relation between the gear rotation and link rotation as
$\theta_d = \phi_d + \theta_d^0$ and $\theta_f = \phi_f + \theta_f^0,$
where $\theta_d^0$ and $\theta_f^0$ are the links' initial angles at $t=0$. The angular velocity of the driver and follower angles can be defined by the link angles as $\dot{\theta}_d = \dot{\phi}_d$ and $\dot{\theta}_f = \dot{\phi}_f$. For continuous cyclic actuation, the driver angle must increase strictly, i.e., $\dot{\theta}_d(t) > 0$ and $\dot{\theta}_f(t) < 0$ due to the engagement constraints. For cyclic gaits with period $T$, the gear rotation should satisfy 
\begin{equation}
     \phi_d(T) - \phi_d(0) = 2\pi \ \text{and} \ \phi_f(T) = \phi_f(0) + K\cdot 2\pi,
\end{equation}
where K is a rational number. In this paper, we focus on the common case of $K=1$, which corresponds to a one-to-one transmission ratio over a gait cycle. We leave it as future work to extend the formulation to arbitrary K.

We express the rotation angle of the follower gear as a function of the rotation angle of the driver gear:
\begin{equation}
    \phi_f = g(\phi_d).
    \label{eq:follower_angle}
\end{equation}
To enforce continuous and monotonic rotation, we represent $g(\phi_d)$ using a \emph{monotone cubic B-spline} with $v$ control points denoted as ${\mathbf{D} \subset \mathbb{R}^{v \times 2}}$, under the constraint $g(2\pi) - g(0) = 2\pi$.
Given the links' initial angles $\theta_d^0$ and $\theta_f^0$, and the function $g(\phi_d)$ in Equation~\eqref{eq:follower_angle}, 
the links' movement can be determined by the driver gear's rotation angle $\phi_d$.
Given the function $g(\phi_d)$, the gear profiles can be computed using the method in~\cite{Xu-2020-Non-CircularGears}. 

Thus, we represent a leg design as $\mathcal{L} = (\mathbf{L}, \mathbf{p}_f, \mathbf{D}, \theta_d^0, \theta_f^0)$, where $\mathbf{L}, \theta_d^0$ and $\theta_f^0$ define the linkage geometry, $\mathbf{p}_f$ specifies the gear base offset, and $\mathbf{D}$ defines the gear profile.

\subsection{Modeling and Actuating LEGO Walkers}
We define a LEGO walker as a combination of a target body frame $\mathcal{B}$ and a set of $N$ legs with the same mechanical design $\mathcal{L}$. The body serves as the central frame to which the legs are attached, and its 3D geometry can be customized to achieve the desired appearance. We represent the body as a 3D mesh.
We define the world origin at the center of mass (COM) of the input body, $+X$ as the walker's forward moving direction, and $+Z$ as the world's vertical up direction.
Each leg mechanism is synthesized and optimized in its local 2D $XY$ plane for kinematic design. After assembly, %
each leg operates in the global $XZ$ plane of the walker according to its mounting configuration. %
The mapping between the $i$-th leg's local frame and the walker's frame is defined by a transformation matrix $\mathbf{M}_i$. To support diverse locomotion configurations, we further allow mirror transformations to mount legs %
in different orientations.

All legs share a common mechanical design $\mathcal{L}$ but may differ in mounting pose and actuation. We index the legs according to their footfall order within a gait cycle, such that leg $i$ is the $i$-th leg to enter the support phase. Each leg is driven by a time-dependent angle trajectory of its driver joint, denoted by $\theta_{d}^i(t)$. To satisfy kinematic constraints and adapt to the leg mechanism $\mathcal{L}$, its mounting configuration $\mathbf{M}_i$, we define the actuation as a function that maps these factors to the driver angle trajectory:
\begin{equation}
\theta_{d}^i(t) = \tau(t; \mathcal{L}, \mathbf{M}_i, i).
\end{equation}
The actuation $\tau(\cdot)$ along with the input parameters should satisfy the kinematics requirements of the stable walking gaits in the next subsection to ensure stable walking.

\subsection{Stable Walking Gaits}
\label{sec:gaits_theory}

\begin{figure}[t]
    \centering
    \includegraphics[width=0.99\columnwidth]{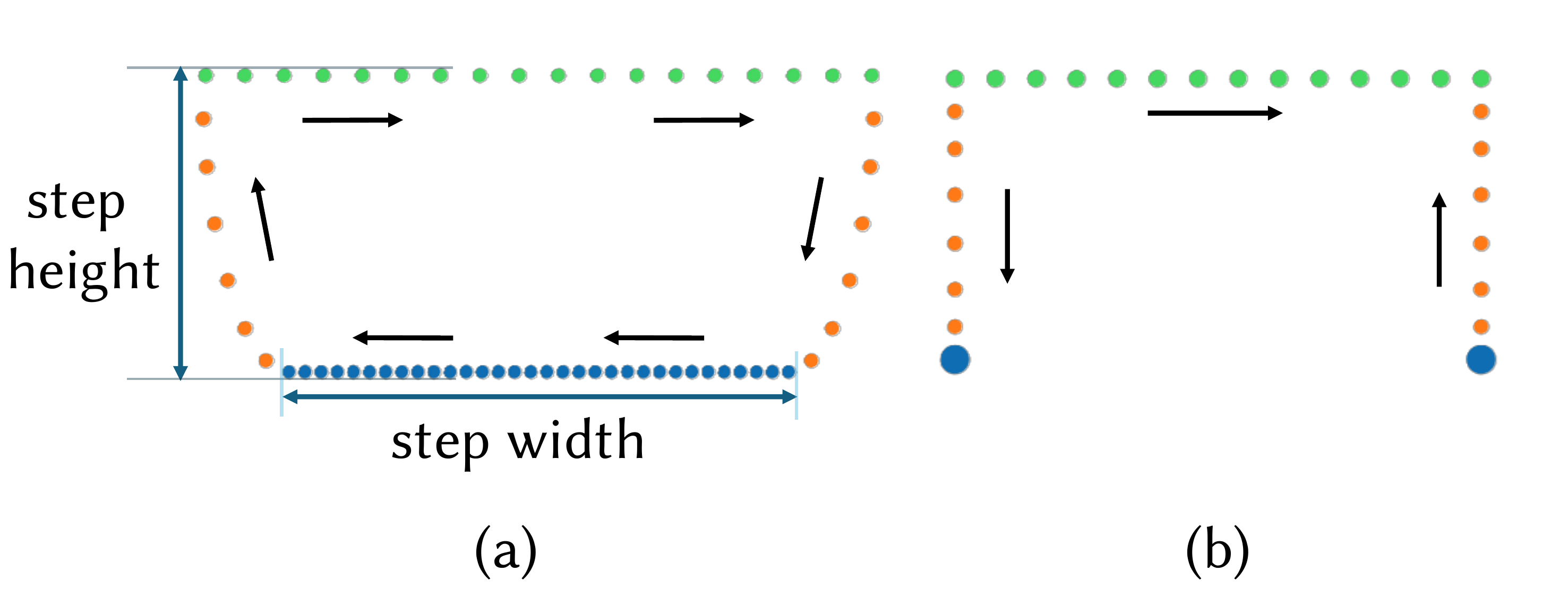}
    \vspace{-5pt}
    \caption{ The foot locus of a leg mechanism. The foot locus is the trajectory traced by the foot over a single gait cycle, consisting of a support phase (in blue) and a transfer phase (in yellow and green). (a) The foot locus when the walker is stationary. (b) The foot locus when the walker is moving.}
    \label{fig:foot_locus}
\end{figure}

We leverage existing statically-stable walking gaits to guide the locomotion of LEGO walkers.
By satisfying requirements of stable walking gaits, the walker can maintain balance during walking without %
dynamic stabilization or complex control strategies.

Stable walking gaits, often referred to as creeping gaits, have been systematically studied in legged locomotion research~\cite{Shigley-1960-Mechanics,Mcghee-1968-Some,Mcghee-1968-Stability}. These gaits are commonly observed in slow-moving animals, maintaining stability by having at least three legs contacting the ground at any time. 
In this kind of gaits, the \emph{foot locus} is the trajectory traced by a foot over a single gait cycle; it consists of a support phase (ground contact to propel the body) and a transfer phase (foot swing to the next contact point). Each foot locus is parameterized by step height $H$, step width $W$, and duty factor $D$, which is the ratio of the support phase duration to the total gait cycle duration. %
Figure~\ref{fig:foot_locus} illustrates a foot locus. 
The body velocity is calculated by $V = \frac{W}{T}$, where $T$ is the gait cycle period.

\begin{figure}[t]
    \centering
    \includegraphics[width=1\columnwidth]{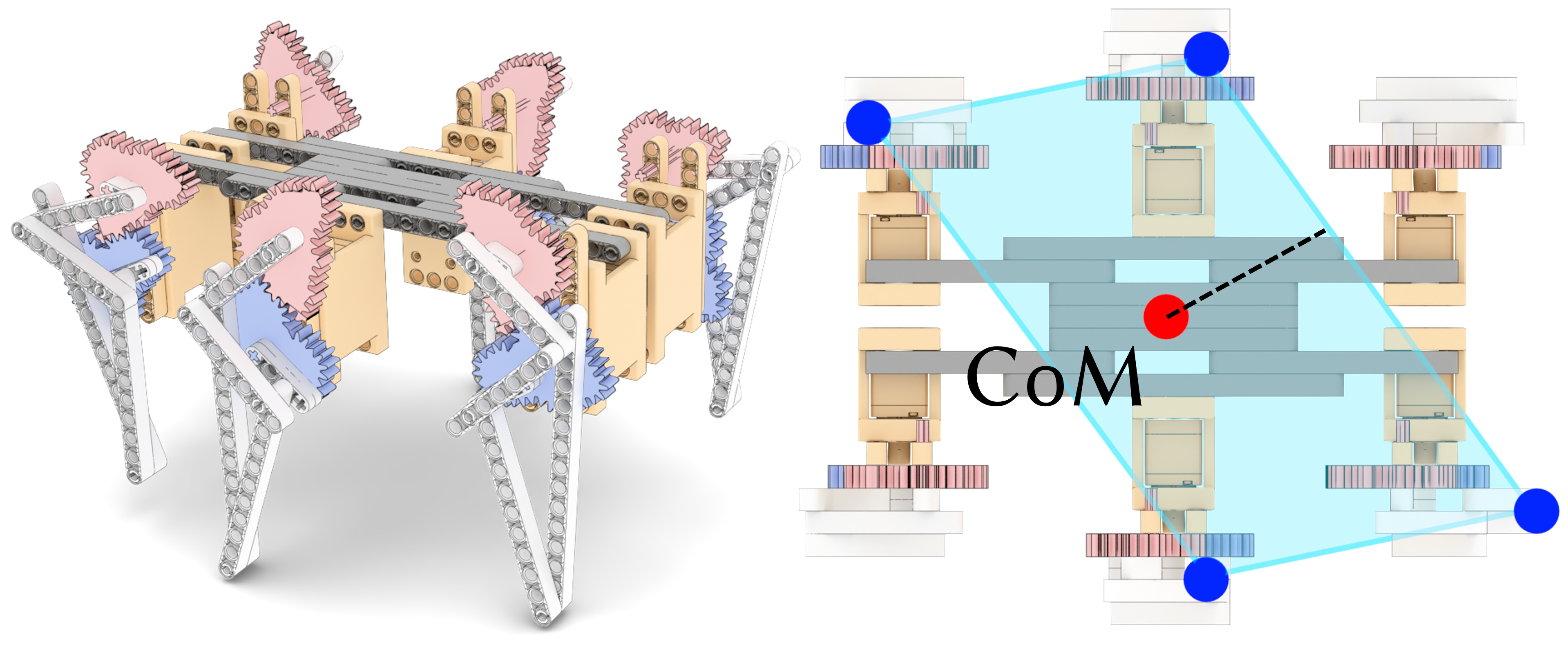}
    \vspace{-5pt}
    \caption{An illustration of the static stability condition. The red dot denotes the projection of the center of mass (COM) onto the ground plane, whereas the blue polygon represents the support polygon formed by the feet currently in contact with the ground (blue dots). The static stability margin is illustrated by the black dashed line.}
    \label{fig:StaticStability}
    \vspace{-10pt}
\end{figure}

For stable walking, the support phase must maintain ground contact, exhibit constant velocity, and have a large enough duty factor. 
The minimum duty factor $\beta_\text{min}$ required to maintain static stability is $\frac{m}{n}$, where \(n\) is the total number of legs and \(m\) is the number of legs that must simultaneously remain in contact with the ground.
For example, $\beta_\text{min}$ is equal to $\frac{3}{4}$ for a quadruped robot.
We adopt the \emph{static stability} criterion, which requires the gravitational projection of the center of mass (COM) to lie strictly within the support polygon formed by the grounded feet. We quantify this using the static stability margin $m(t)$, defined as the distance from the COM projection to the support polygon boundary at time $t$ (Figure~\ref{fig:StaticStability}). A walking gait is considered stable if $m(t) > 0$ for any $t \in [0, T]$. 

%% file: sec5-design.tex
\section{Designing LEGO Walkers}
\label{sec:design}
In this section, we formulate the design of LEGO walkers as an optimization problem and define its search space, objective functions, and constraints. 
Then, we present a method to solve the optimization problem.

\subsection{Optimization Formulation}

\paragraph*{Search space.} 
The search space is hybrid, consisting of the parameters of the leg mechanisms denoted by $\mathcal{L}$, and the parameters for the walker including the matrix of each leg $\mathbf{M}_i$ and the actuation control for each leg $\theta_d^i(t)$. 
In addition, we search for two auxiliary parameters, $s_l$ and $s_r$, which define the support phase by specifying its start time $s_l$ and end times $s_r$ within a gait cycle.
Note that the actuation control $\theta_d^i(t)$ determines the temporal aspect of the foot locus, whereas the other parameters determine the spatial (shape) aspect of the foot locus.

\paragraph*{Objective functions.} 
Our objective is to design a fabricable LEGO walker that generates a stable walking gait according to target specifications. 
To this end, we define the following objective functions:
\begin{itemize}[leftmargin=*]

    \item 
    {\em Walking stability.} \ 
    We maximize the the minium stability margin during the gait cycle, to ensure stable walking:
    \begin{equation}
        E_s = \min_{t \in [0, T]} m(t). 
        \label{eq:stability_energy}
    \end{equation}

    \item {\em Gait approximation.} 
    We encourage the foot locus to approximate the target 
    trajectory by minimizing the discrete Fréchet distance between the foot trajectory $\mathbf{P}_e(t)$ and the target trajectory $\mathbf{P}_\text{target}(u)$:
    \begin{equation}
        E_e = d_F(\mathbf{P}_e(t), \mathbf{P}_\text{target}(u)),
        \label{eq:gaits_approximation}
    \end{equation}
    where $d_F$ denotes the discrete Fréchet distance, and $u$ is the parameter of the target trajectory curve $\mathbf{P}_\text{target}(u)$.
    
\end{itemize}

\paragraph*{Constraints.} 
The foot locus must satisfy the constraints of stable walking gaits, which include:
\begin{itemize}[leftmargin=*]
    \item 
    {\em Singularity-free motion.} \  
    To avoid singularities, the mechanism must keep a safe distance from singular configurations throughout the gait cycle:
    \begin{equation}
    \left| \| \mathbf{q}_d(t) - \mathbf{q}_f(t) \| - (l_c^1 \pm l_c^2) \right| > \epsilon, \quad \forall t \in [0, T],
    \label{eq:singularity_constraint}
    \end{equation}
    where $\epsilon$ denotes a safety margin. We set $\epsilon=0.2$ in our experiments.

    \vspace*{1mm}
    \item 
    {\em Collision-free motion.} \ 
    Each leg should avoid collision with the walker body and other legs:
    \begin{equation}
        \begin{aligned}
        d\!\left(\mathcal{X}_i(t), \mathcal{X}_j(t)\right) \ge \delta
        &\quad \forall t\in[0,T], \\
        \forall \mathcal{X}_i \neq \mathcal{X}_j \in
        &\{\mathcal{B},\mathcal{G}_{1},\dots,\mathcal{G}_{n}\}.
        \end{aligned}
        \label{eq:collision}
    \end{equation}
    where $\mathcal{G}_{i}$ denotes the geometry of the $i$-th leg defined by the leg parameters $\mathcal{L}$ and transformation $\mathbf{M}_i$, $d(\cdot,\cdot)$ denotes the minimum Euclidean distance between two geometries, and $\delta$ is a prescribed clearance threshold.

    \vspace*{1mm}
    \item 
    {\em Ground contact.} \
    Each foot, during the support phase, must remain in contact with the ground:
    \begin{equation}
        [\mathbf{P}_e(t)]_y = 0, \quad t \in [s_l, s_r].
        \label{eq:ground_contact}
    \end{equation}

    \item 
    {\em Duty factor.} \
    For static stability, the duty factor must exceed a %
    threshold $\beta_\text{min}$,
    which is determined by the number of simultaneously supporting legs relative to the total number of legs:
    \begin{equation}
        \frac{s_r - s_l}{T} > \beta_\text{min}.
        \label{eq:duty_factor}
    \end{equation}

    \item 
    {\em Constant contact velocity.} \ 
    The foot velocity during the support phase should remain constant:
    \begin{equation}
        \dot{\mathbf{P}}_e(t) = \mathbf{V}, \quad \forall t \in [s_l, s_r].
        \label{eq:constant_velocity}
    \end{equation}

    \item 
    {\em Clearance.} \ 
    Each foot during the transfer phase must maintain a minimum height $H_\text{min}$ to avoid ground collision:
    \begin{equation}
        [\mathbf{P}_e(t)]_y \ge H_\text{min}, \quad \forall t\in [0,T]\setminus[s_l,s_r]. 
        \label{eq:clearance}
    \end{equation}

    \item 
    {\em Fabricability.} \ 
    The non-circular gear profiles must satisfy fabrication constraints: the instantaneous transmission ratio should lie within a reasonable range to ensure smooth motion: 
    \begin{equation}
        r_\text{min} \le \frac{\dot{\theta}_d(t)}{\dot{\theta}_f(t)} \le r_\text{max}, \quad \forall t \in [0, T].
        \label{eq:fabricability}
    \end{equation}
    We set $r_\text{min} = 0.3$ and $r_\text{max} = 3.0$ in our experiments.
\end{itemize}

\subsection{Optimization Solver}
\label{sec:optimization_solver}
Our key observation for the optimization is that (i) the constraints and objectives can be categorized into spatial and spatio-temporal groups, and (ii) the spatial and temporal requirements can be addressed separately through mechanical design and electronic control, respectively. Based on this spatio-temporal decoupling, we first design the leg mechanisms to generate the target foot trajectories to satisfy the spatial requirements, and then optimize the actuation control to fulfill the temporal requirements.

We categorize the objective functions and constraints into spatial and spatio-temporal groups. The energy terms in Equations~\eqref{eq:stability_energy} and~\eqref{eq:gaits_approximation} are spatial objectives, as they depend directly on the geometric shape of the foot locus and can be optimized through the leg mechanism design. Similarly, the constraints in Equations~\eqref{eq:singularity_constraint},~\eqref{eq:ground_contact},~\eqref{eq:clearance}, and ~\eqref{eq:fabricability} are purely spatial, and can likewise be satisfied via the mechanism design.
The constraints in Equations~\eqref{eq:collision},~\eqref{eq:duty_factor}, and~\eqref{eq:constant_velocity} are spatio-temporal, as they involve both geometric properties and timing requirements. These constraints are jointly governed by the leg mechanism design and the actuation. According to Section~\ref{sec:gaits_theory}, when satisfying both the spatial and temporal requirements, the walker can walk stably in practice. 
Based on this categorization, we propose a three-stage method to solve the optimization problem, where the first two stages determine the leg geometry and layout to satisfy the spatial targets and the third stage determines the leg actuation to satisfy the temporal targets.

%% file: sec6-result.tex
\begin{figure*}[!t]
    \centering
    \includegraphics[width=0.92\textwidth]{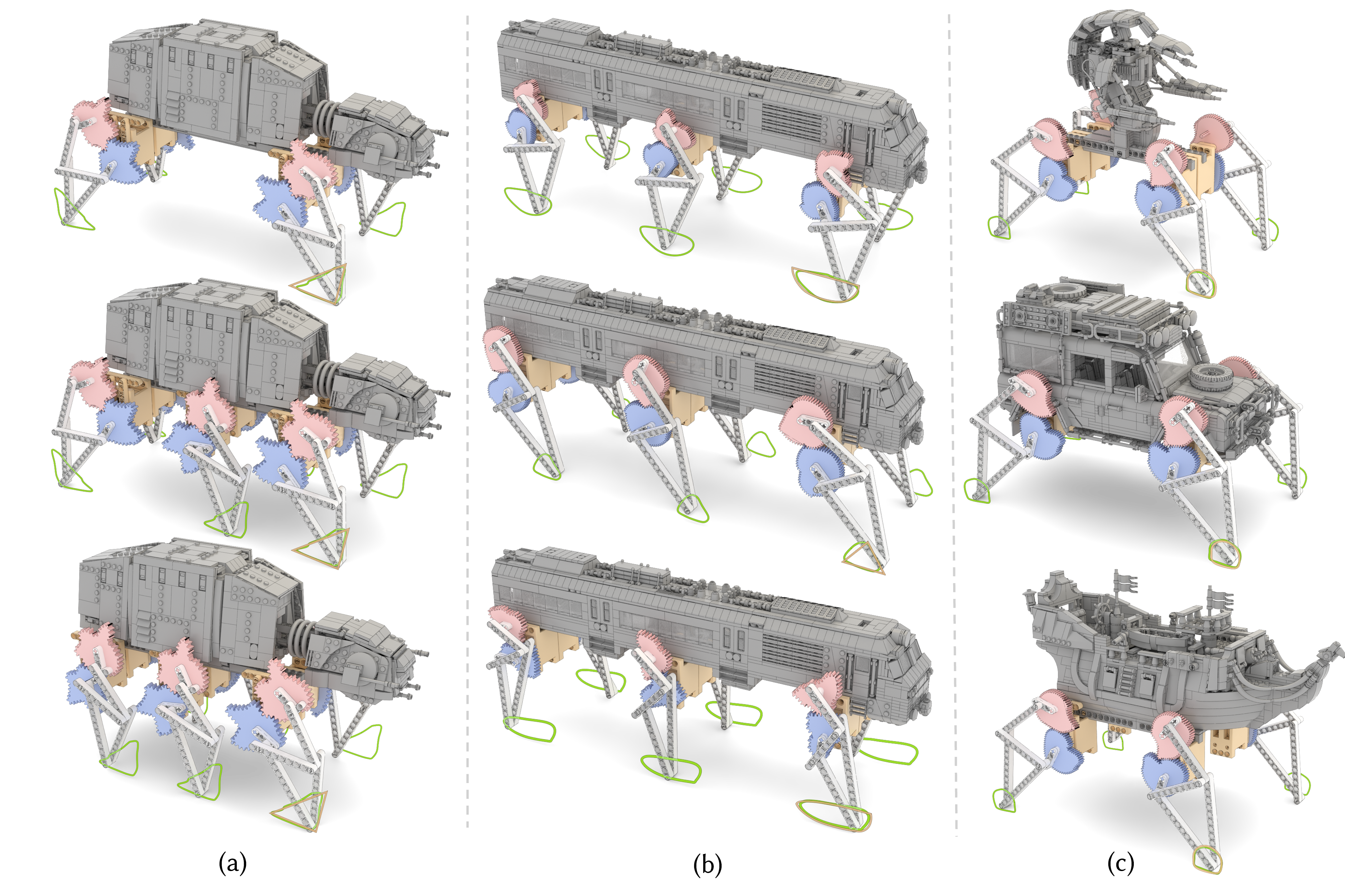}
    \vspace{-10pt}
    \caption{Our approach allows designing a variety of LEGO walkers.
    For each result, the target foot trajectory is shown in yellow at the walker's right front leg, while the actual foot trajectories are shown in green for all the legs.
    (a) We vary the number of legs and the leg layout, while fixing the body frame and the target foot trajectory.
    (b) We vary the target foot trajectory, while fixing the body frame and the number of legs.
    (c) We vary the body frame, while fixing the number of legs and the target foot trajectory.}
    \label{fig:body_variation}
    \vspace{-10pt}
\end{figure*}

\begin{figure}[!t]
    \centering
    \includegraphics[width=0.9\columnwidth]{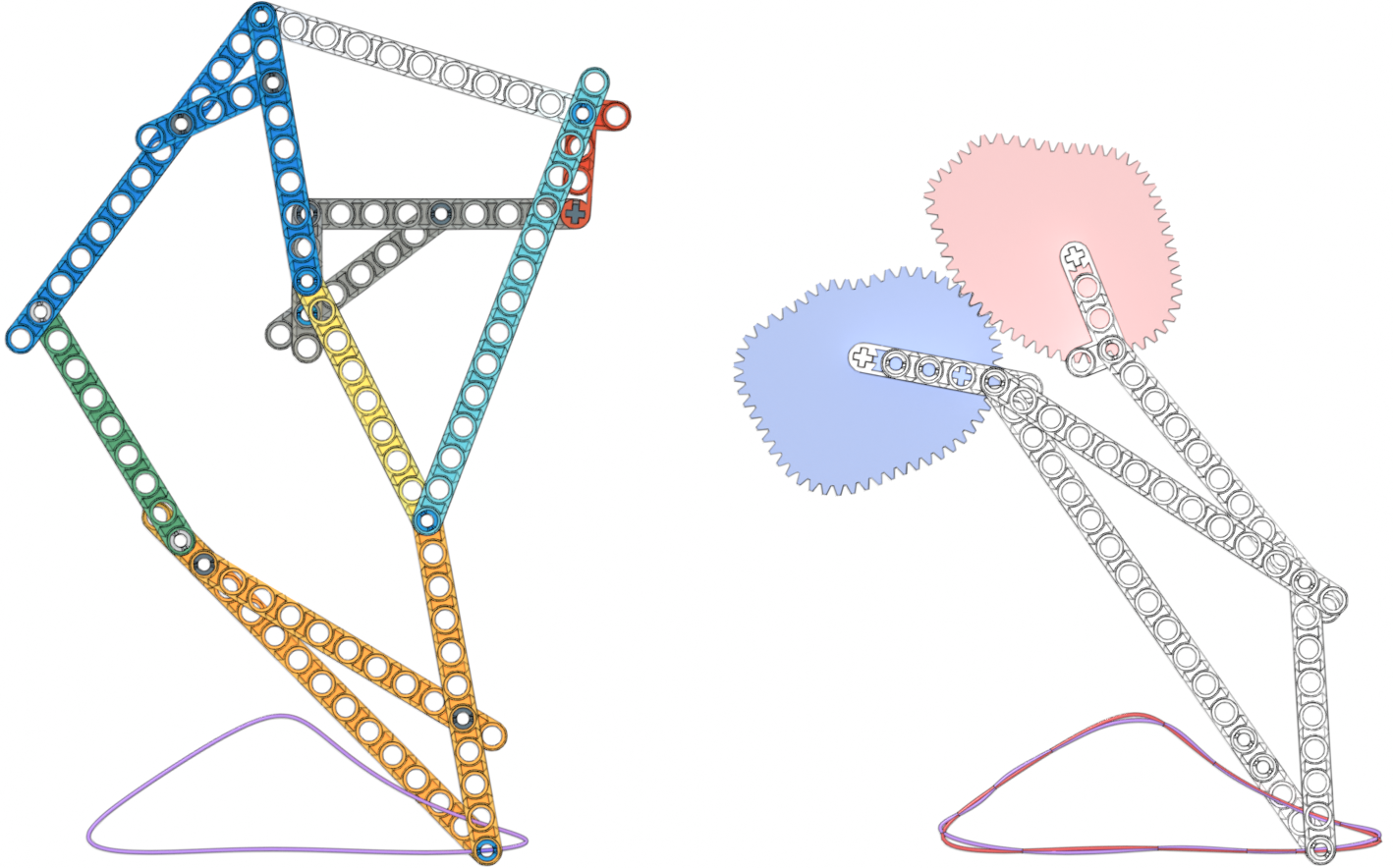}
    \caption{
    Taking the foot trajectory of (left) the Jansen linkage as the target curve, (right) our designed leg mechanism generates a foot trajectory that closely matches the target.
    Compared with the Jansen linkag, our leg mechanism has fewer parts, a simpler mechanical structure, and a smaller size.
    }
    \label{fig:jasen_compare}
    \vspace{-15pt}
\end{figure}

\begin{figure}[!t]
    \centering
    \includegraphics[width=0.95\columnwidth]{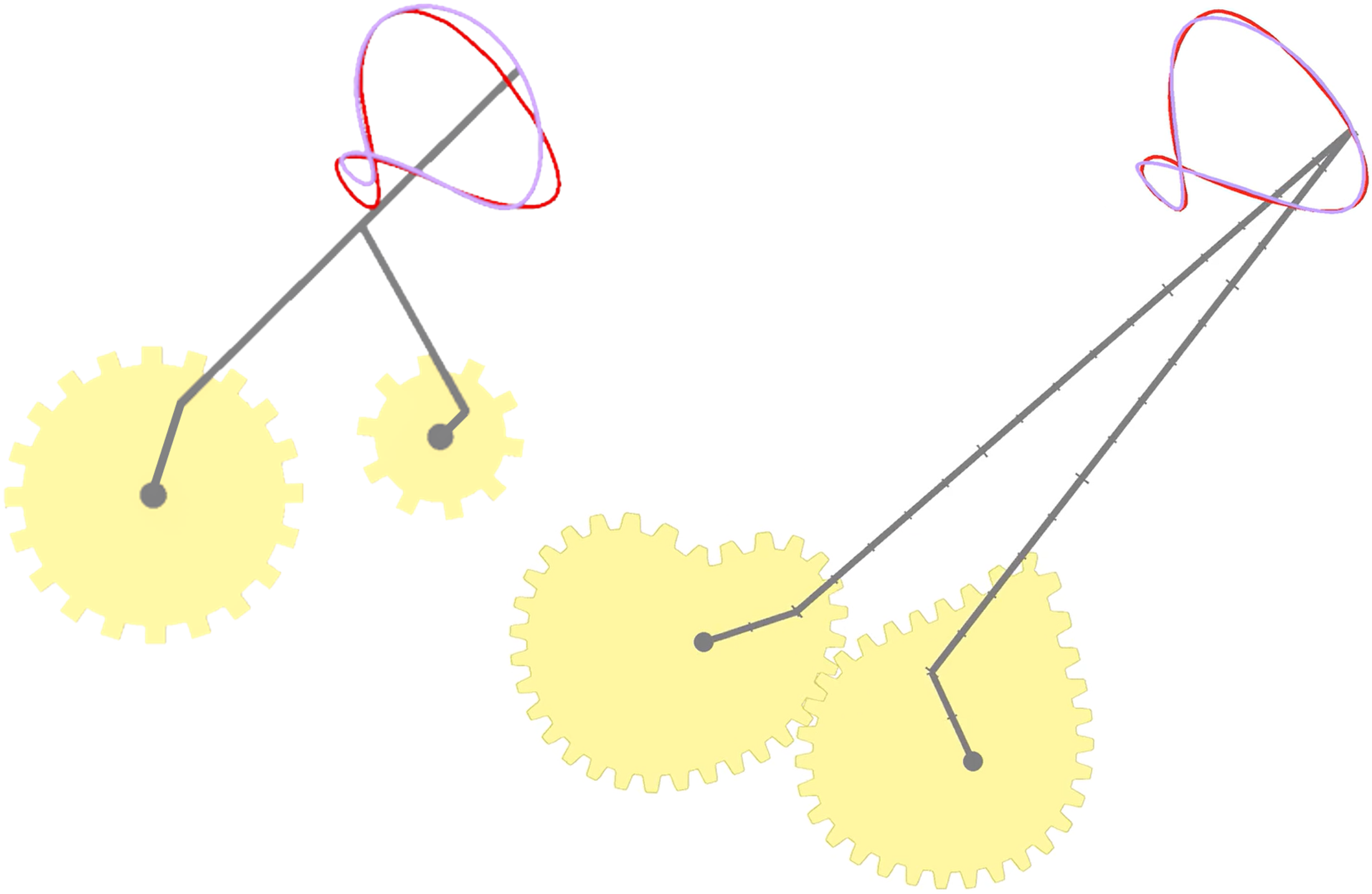}
    \caption{
    Given the same target curve (in red color), (right) our leg mechanism generates a curve that more closely matches the target curve than (left) that of Coros et al.~\cite{Coros-2013-MechanicalCharacter}, due to the larger design space enabled by the non-circular gears.
    }
    \label{fig:comparision}
    \vspace{-15pt}
\end{figure}

\section{Results}
\label{sec:result}
We implemented our approach in Python on a desktop computer with a 4.3GHz CPU and 32GB memory, %
and employed PyTorch~\cite{Paszke-2019-Pytorch} for solving the optimization in Section~\ref{sec:design}. 
All the LEGO walker results shown in the paper are optimized by our approach in less than 30 minutes.
These LEGO walkers have four or six legs, where a larger number of legs typically lead to higher walking stability (i.e., larger $E_s$ in Equation~\eqref{eq:stability_energy}).

\paragraph*{Virtual results.}
Figure~\ref{fig:body_variation} shows a variety of LEGO walkers designed by our approach.
In particular, our approach allows designing LEGO walkers with different numbers of legs and leg layouts (Figure~\ref{fig:body_variation}(left)), different target foot trajectories (Figure~\ref{fig:body_variation}(middle)), and different body frames (Figure~\ref{fig:body_variation}(right)). 
For each of these results, our approach computes the geometry, layout, and control of the leg mechanisms to achieve the target walking gait and match the given body frame.
Please refer to the accompanying video for the simulation of these walkers' locomotion.

\paragraph*{Comparison with a LEGO-discretized Jansen linkage.}
We compare our leg mechanism against a LEGO-discretized version of the well-known Jansen linkage to show that our leg mechanism is able to perform the same function yet has a simpler mechanical structure. 
Taking the foot trajectory of the Jansen linkage as the input, we run our optimization in Section~\ref{sec:design} to design a leg mechanism for generating the trajectory curve.
As Figure~\ref{fig:jasen_compare} shows, our designed leg mechanism is able to produce a foot trajectory that closely matches that of the Jansen linkage.  Yet, our mechanism has a considerably more compact and simpler mechanical structure. Specifically, our mechanism consists of only 10 link parts, 8 pins, and two gears, with a total beam length of \(76L\), where $L$ denotes one unit of LEGO Technic beam length.
In constrast, the discretized Jansen linkage requires 14 link parts, 14 pins, with a total beam length of \(134L\).

\begin{figure*}[!t]
    \centering
    \includegraphics[width=0.98\textwidth]{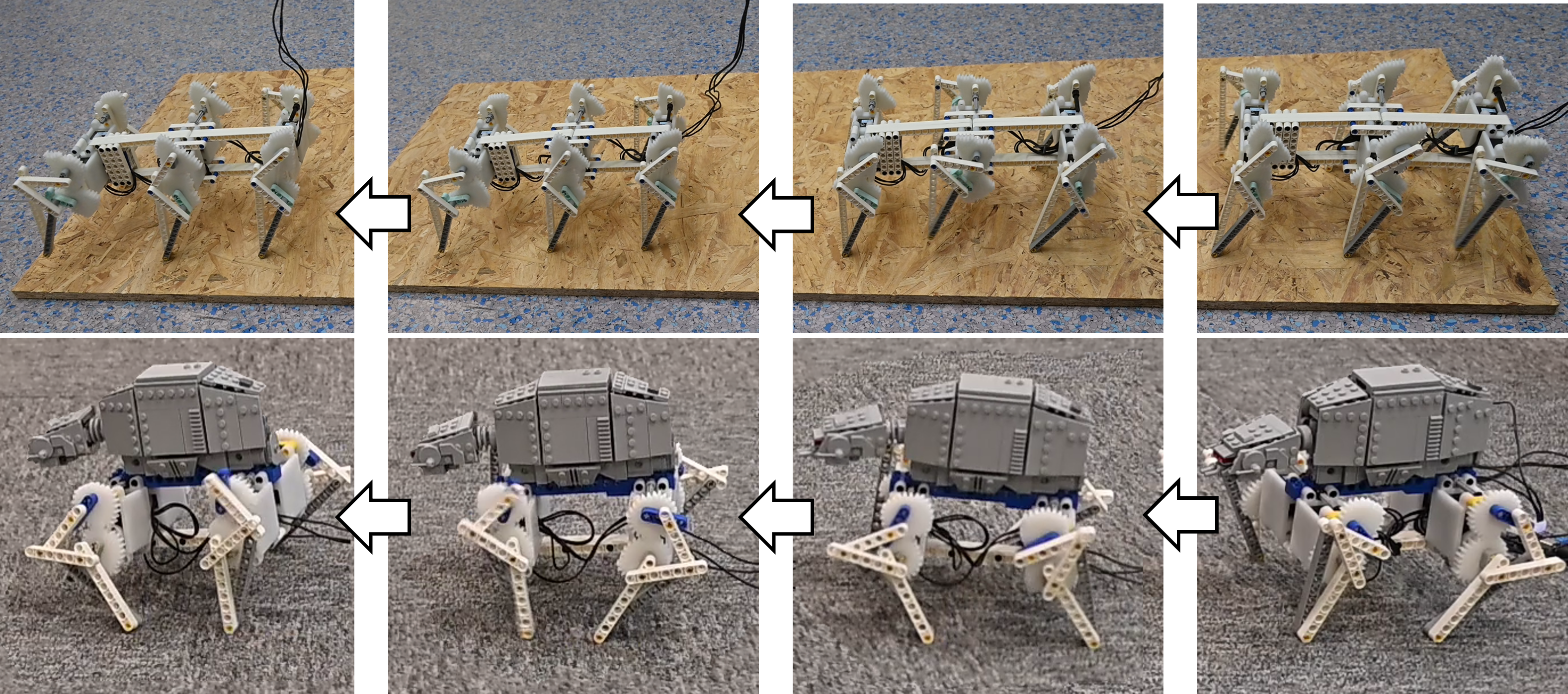}
    \caption{Key frames of two LEGO walkers designed by our approach and fabricated using LEGO Technic and 3D printing.
    (Top) A hexapod LEGO walker without a body frame performs locomotion on uneven terrain. 
    (Bottom) A quadruped LEGO walker with a Sci-Fi body frame performs locomotion on flat terrain.
    Please refer to the accompanying video for the full walking locomotion.}
    \label{fig:GaitsIllustration}
    \vspace{-10pt}
\end{figure*}

\begin{figure}[!t]
    \centering
    \includegraphics[width=0.82\columnwidth]{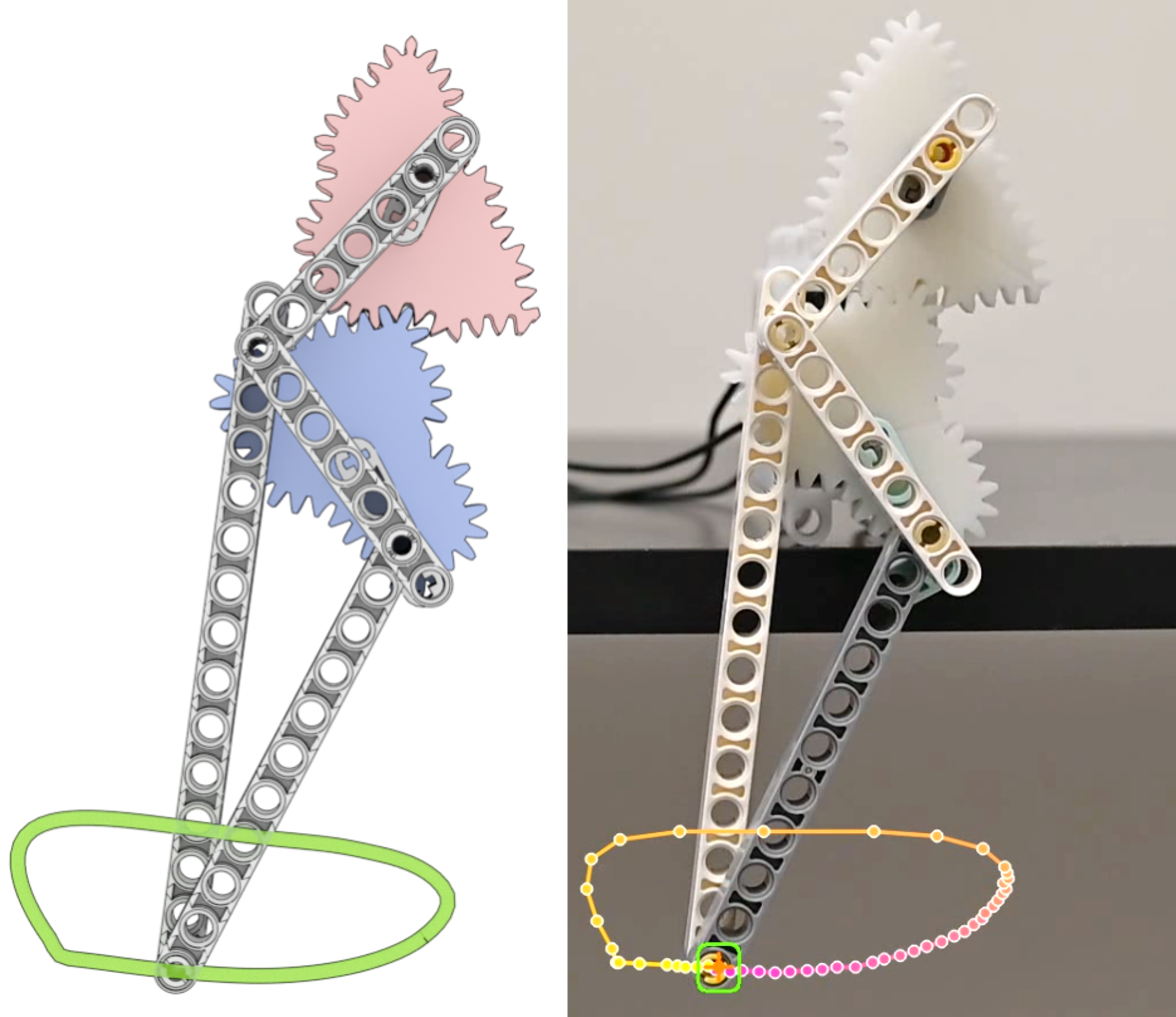}
    \vspace{-5pt}
    \caption{Evaluating the kinematic performance of a leg mechanism designed by our approach.
    (Left) Our designed leg mechanism, where the simulated foot trajectory is shown in green.
    (Right) Corresponding view of our LEGO-built prototype, where the actual foot trajectory is tracked on video images and visualized in orange.}
    \label{fig:curve_compare}
    \vspace{-5pt}
\end{figure}

\paragraph*{Comparison with [Coros et al. 2013].}
We compare our leg mechanism with that proposed by Coros et al.~\cite{Coros-2013-MechanicalCharacter} to demonstrate its performance in generating a user-prescribed 2D path. 
Both mechanisms consist of a five-bar linkage. 
The key difference is that our mechanism uses a pair of non-circular gears to couple the two linkage inputs, while the mechanism of Coros et al.~\cite{Coros-2013-MechanicalCharacter} uses circular gears. 
Given the same target curve, we run our optimization in Section~\ref{sec:design} to design a leg mechanism for generating the target curve.
Figure~\ref{fig:comparision} shows that our mechanism generates a curve that more closely matches the target curve than that of Coros et al.~\cite{Coros-2013-MechanicalCharacter}, achieving a normalized Fréchet distance of 0.073 compared with 0.255.
We attribute this improved performance to the larger design space of our mechanism: the non-circular gears provide continuous pitch-curve design freedom, whereas the circular gears are characterized by only a single gear ratio.

\begin{figure}[!t]
    \centering
    \includegraphics[width=0.8\columnwidth]{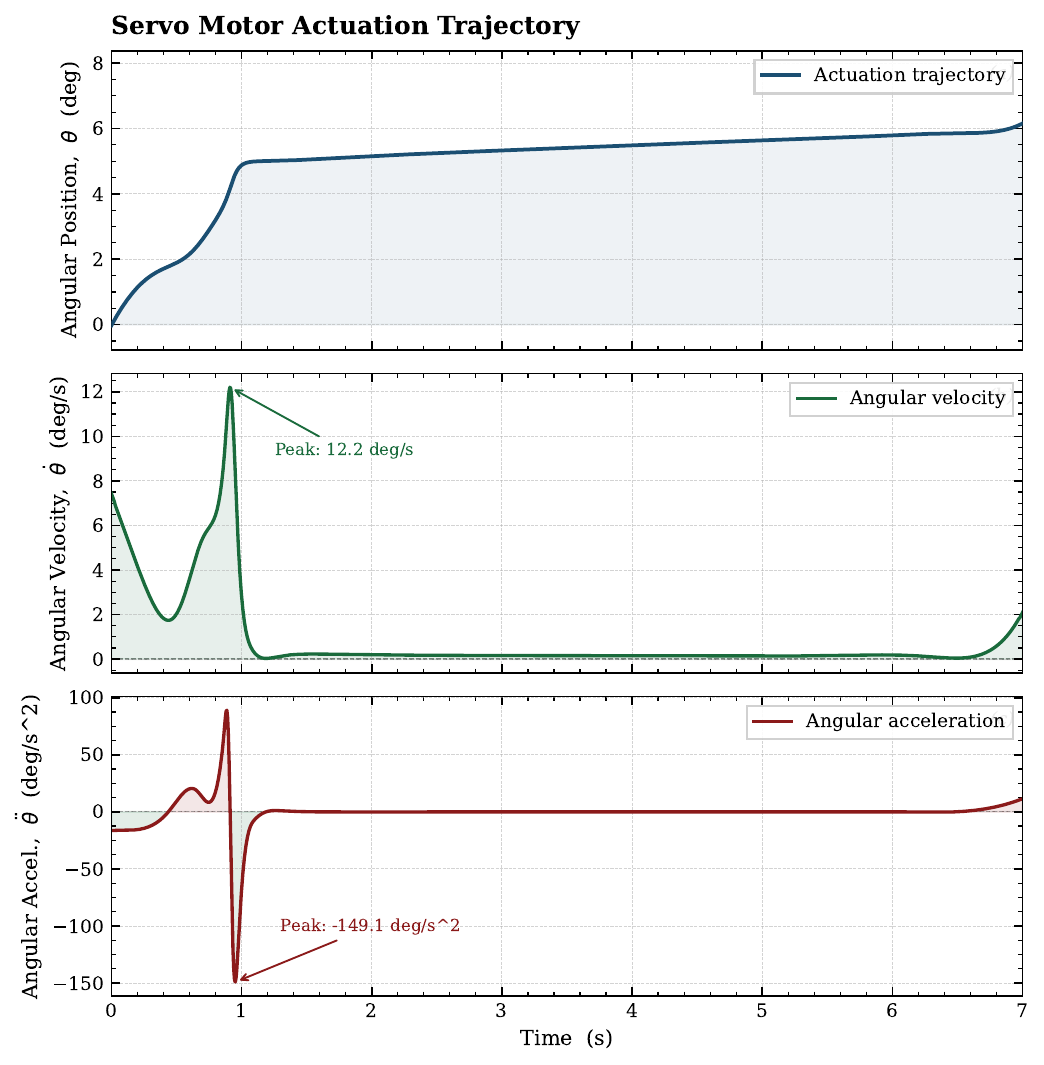}
    \vspace{-5pt}
    \caption{Control sequence analysis on one of the leg mechanisms in our hexapod LEGO walker shown in Figure~\ref{fig:GaitsIllustration}.}
    \label{fig:control_sequence}
    \vspace{-10pt}
\end{figure}

\paragraph*{Evaluating our leg mechanism.} 
To evaluate the kinematic performance of our leg mechanism, we used our approach to design a leg mechanism (see Figure~\ref{fig:curve_compare} (left)) and fabricated it using LEGO Technic parts and non-circular gears 3D printed by the Lite SPSS600 printer with UV-curable resin material (Figure~\ref{fig:curve_compare} (right)).
We compare the simulated foot trajectory with the one tracked from the physical prototype. 
The simulated trajectory exhibits clear separation between support and transfer phases, with flat and stable ground contact segments. 
The tracked trajectory on real-world video images closely aligns with the simulated trajectory, manifesting that our kinematic model and optimization accurately predict the physical behavior despite simplifications in modeling the LEGO joints and tolerances.

\paragraph*{Physical fabrications.}
Figure~\ref{fig:GaitsIllustration} shows two physically-fabricated %
LEGO walkers designed by our approach. 
Each row in the figure presents four key frames of a gait cycle. 
For both the hexapod (top row) and quadruped (bottom row) walkers, the designed mechanisms are able to deliver coordinated leg motions that yield continuous forward locomotion. 
In addition, Figure~\ref{fig:control_sequence} shows one leg's actuation control for the hexapod walker. 
The actuation of the leg is not a simple linear sequence but a rather complex pattern that requires relatively precise control to achieve the desired foot trajectory and timing. Nevertheless, the required actuation remains moderate, with a peak angular acceleration of only $149.1^\circ/\mathrm{s}^2$. 
These physical realizations demonstrate that our designed walkers can be reliably driven by the physical system while effectively synchronizing the leg motions to deliver stable walking.

%% file: sec7-conclusion.tex
\section{Conclusion}
\label{sec:conclusion}

This paper presents a new class of LEGO walkers that decouples the trajectory and timing control of the legs, enabling the walkers to have a relatively simple morphology and controller. A computational approach is introduced for designing LEGO walkers with varying walking gaits and stable locomotion. To do so, we formulate the co-design of the LEGO walkers' morphology and control as an optimization problem and propose a method to effectively explore the vast search space. We demonstrate that our approach can design LEGO walkers with a variety of walking gaits and appearance. Also, we fabricated two of these walkers and demonstrated their stable locomotion and walking gaits in the real world.
\paragraph*{Limitations and future work.}
First, our current approach is limited to statically stable creeping gaits. Although this formulation avoids the need for expensive sensors and complex locomotion control, it restricts the range of achievable gaits. In the future, we plan to extend our approach to dynamically stable gaits and explore locomotion over more complex terrains, such as stairs and rough ground.
Second, the legs of our LEGO walkers currently share the same mechanism design. Allowing different leg mechanisms in a single walker could enrich the diversity of walking gaits.